%% file: main.tex
\documentclass[letterpaper, 10 pt, conference]{ieeeconf}  

\IEEEoverridecommandlockouts                              
\usepackage{todonotes}
\usepackage{graphicx} 
\usepackage{times} 
\usepackage{amsmath} 
\usepackage{amssymb}  
\usepackage{verbatim} 
\usepackage{cite}
\usepackage{subfig}
\usepackage[hyphens]{url}
\usepackage[ruled,vlined]{algorithm2e}
\usepackage{float}
\usepackage{tabulary}
\usepackage{tabularx}
\usepackage{lipsum,amsmath,multicol}
\usepackage{tensor}
\usepackage{multirow}
\usepackage{booktabs}
\usepackage{flushend}
\usepackage[flushleft]{threeparttable}
\usepackage{soul}
\input{./extern/pic-common}

\usepackage{color}

\definecolor{CommentRed}{rgb}{0.7,0,0}
\definecolor{CommentBlue}{rgb}{0,0,0.7}
\definecolor{CommentDG}{rgb}{0,0.6,0}
\definecolor{CommentPink}{rgb}{0.7,0,0.7}

\usepackage{xcolor}

\makeatletter
\newenvironment{myalign*}{%
  \setlength{\mathindent}{0pt}%
  \setlength{\abovedisplayskip}{-\baselineskip}%
  \setlength{\abovedisplayshortskip}{\abovedisplayskip}%
  \start@align\@ne\st@rredtrue\m@ne
}%
{\endalign}
\makeatother

\title{\LARGE \bf
An Overview of Perception Methods for Horticultural Robots: From Pollination to Harvest
}

\author{Ho Seok Ahn$^{1*}$, Feras Dayoub$^{2}$, Marija Popovi\'{c}$^{3}$, Bruce A. MacDonald$^{1}$, Roland Siegwart$^{3}$, and Inkyu Sa$^{3}$
\thanks{Ho Seok Ahn$^{1}$ and Bruce A. MacDonald$^{1}$ are with the Department of Electrical and Computer Engineering, CARES, University of Auckland, 368 Khyber Pass Rd Newmarket Auckland 1023, New Zealand. 
        {\tt\small \{hs.ahn, b.macdonald\}@auckland.ac.nz}}%
\thanks{Feras Dayoub$^{2}$ is with the Australian Centre for Robotic Vision at Queensland University of Technology (QUT).
        {\tt\small feras.dayoub@qut.edu.au}}%
\thanks{Marija Popovi\'{c}$^{3}$, Roland Siegwart$^{3}$, and Inkyu Sa$^{3}$ are with Department of Mechanical and Process Engineering, Institute of Robotics and Intelligent Systems, Autonomous Systems Lab., ETH Zurich, Switzerland.
        {\tt\small \{mpopovic, rsiegwart, sain\}@ethz.ch}}%
\thanks{*The corresponding author.}%
}

\begin{document}

\maketitle

\thispagestyle{empty}
\pagestyle{empty}





\begin{abstract}

Horticultural enterprises are becoming more sophisticated as the range of the crops they target expands. Requirements for enhanced efficiency and productivity have driven the demand for automating on-field operations.
However, various problems remain yet to be solved for their reliable, safe deployment in real-world scenarios.
This paper examines major research trends and current challenges in horticultural robotics.
Specifically, our work focuses on sensing and perception in the three main horticultural procedures: pollination, yield estimation, and harvesting.
For each task, we expose major issues arising from the unstructured, cluttered, and rugged nature of field environments, including variable lighting conditions and difficulties in fruit-specific detection, and highlight promising contemporary studies. 
Based on our survey, we address new perspectives and discuss future development trends.

\end{abstract}


\input{./contribs/Introduction2.tex}
\input{./contribs/SensingChallenges.tex}
\input{./contribs/Pollination2.tex}

\input{./contribs/YieldEstimation2.tex}

\input{./contribs/Harvesting2.tex}

\input{./contribs/Conclusions2.tex}
\input{./contribs/Acknowledgement.tex}

\bibliographystyle{ieeetr}
\bibliography{bibs/RAL2018}

\end{document}

%% file: extern/pic-common.tex
\def\fps@figure{htp}
\def\fps@table{htp}

\newcommand{\bi}{\begin{itemize}}
\newcommand{\ei}{\end{itemize}}

\newcommand{\bfig}{\begin{figure}}
\newcommand{\efig}{\end{figure}}

\newcommand{\benum}{\begin{enumerate}}
\newcommand{\eenum}{\end{enumerate}}

\newcommand{\be}{\begin{equation}}
\newcommand{\ee}{\end{equation}}

\newcommand{\ba}{\begin{eqnarray}}
\newcommand{\ea}{\end{eqnarray}}

\newcommand{\etal}{{et al.}}

\newcommand{\unit}[1]{\mbox{$\rm \,#1$}}

%% file: contribs/Introduction2.tex
\section{Introduction} \label{sec:intro2}


The horticultural industry faces increasing pressure as demands for high-quality food, low-cost production, and environmental sustainability grow. To cater for these requirements, a top priority in this field is optimizing methods applied at each stage of the horticultural process. However, on-field procedures still rely heavily on manual tasks, which are arduous and expensive. In the past few decades, robotic technologies have emerged as a flexible, cost-efficient alternative for overcoming these issues \cite{Bechar2017}.

\begin{figure}[t]
\begin{center}
\includegraphics[width=0.96\columnwidth]{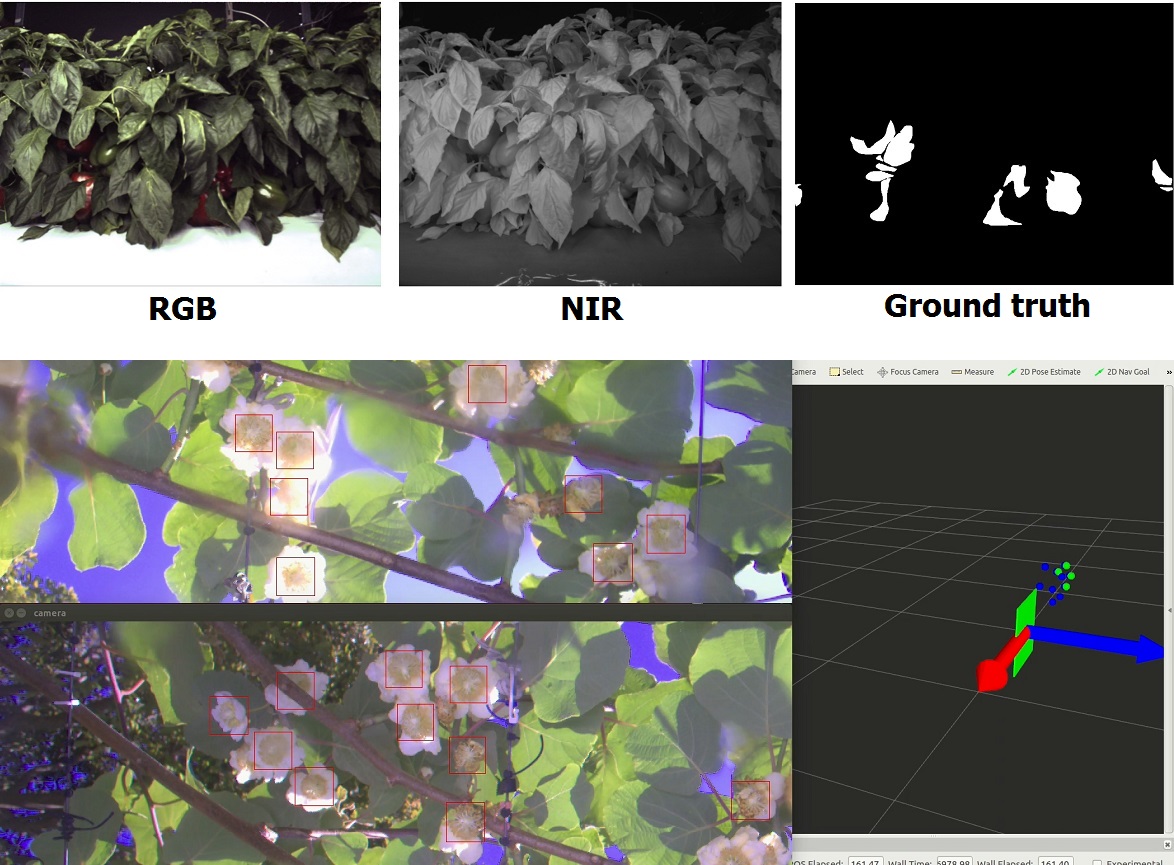}
\end{center}
   \vspace{-3mm}
   \caption{Challenges in perception for horticultural applications. (Top) Red and green sweet peppers are difficult to identify in occluded conditions, even for humans. (Bottom) Matching targets detected in two camera images and finding their 3D positions is hard when perception aliasing occurs.}
    \label{fig:challenge}
    \vspace{-6mm}
\end{figure}

To fully exploit the potential of automated field production techniques, several challenges remain to be addressed. Robots in orchard environments must tackle issues such as the presence of uncontrolled plant growth, weather exposure, as well as the slope, softness, cluttered and undulating nature of the transversed terrain. In addition, there are perceptual difficulties in cluttered environments due to variable illumination conditions and occlusions, as depicted in Fig.~\ref{fig:challenge}.

This paper summarizes recent developments and research in robotics for horticultural applications. We structure our discussion based on three main procedures in the general horticultural process, which correspond to the key stages of plant growth: (1) pollination, (2) yield estimation, and (3) harvesting. Specifically, this paper focuses on robotic perception methods, which are a core requirement for systems operating in all three areas, and an actively researched field of study. For each step, we examine major challenges and outline potential areas for future work. The following provides a brief overview to motivate our study.

\vspace{4mm}



\textbf{Pollination}: Bees are major pollinators in traditional horticulture \cite{Lofaro2017Bee}. However, their numbers are diminishing rapidly due to colony collapse disorder, pesticides, and invasive mites, as well as climate change and inconsistent hive performance \cite{Goulson2015}. Studies conducted between 2015 and 2016 report a total annual colony loss of 40.5\% \cite{Kulhanek2017} in the United States. This leads to a decrease in crop quality and quantity, causing farm owners to hire employees for seasonal hand pollination \cite{Giang2017Vanilla}, which is labor-intensive. To address these issues, robotic systems are being developed to spray pollen on flowers \cite{Abutalipov2016}. 
An important consideration is producing fruit with uniform size and quality to raise their value. 


\textbf{Yield estimation}: Crop production estimates provide valuable information for cultivators to plan and allocate resources for harvest and post-harvest activities. Currently, this process is performed by manual visual inspection of a sparse sample of crops, which is not only labour-intensive and expensive, but also inaccurate, depending on the number of counts taken, and with sometimes destructive outcomes. For this task, automated solutions have also been proposed as an alternative \cite{Nuske2011YieldEI}. Here, a key aspect is designing systems able to operate in unstructured and cluttered environments \cite{Bechar2017}.

\textbf{Harvesting}: The final horticultural procedure performed on-field, harvesting, usually incurs high labor cost due to its repetitive and monotonous nature. Autonomous harvesters have been proposed as a viable replacement which can also procure relatively high-quality products \cite{lehnert2017autonomous}. However, deployment in real orchards requires complex vision techniques able to handle a wide range of perceptual conditions.

This paper is organized as follows. In Section~\ref{sec:SensingChallenges}, we examine the main sensing challenges in horticultural environments. Section~\ref{sec:Pollination2} discusses flower detection and recognition methods for pollination, while Sections~\ref{sec:Yield2} and~\ref{sec:harvesting2} discuss perception challenges for automated yield estimation and harvesting, respectively. Concluding remarks including directions for further research are given in Section~\ref{sec:conclusion2}.

%% file: contribs/SensingChallenges.tex
\section{Sensing Challenges in Horticultural Environments}\label{sec:SensingChallenges}

Our survey focuses on two contemporary technologies as methods of addressing major challenges in developing robots for pollination and harvesting: sensing and perception. 
Successfully detecting crops and their parts plays a crucial role in horticulture, as these processes lay the groundwork for subsequent operations, such as selective spraying or weeding, obstacle avoidance, and crop picking and placing. Recent developments have enabled harvesting and scouting robots to deploy lighter, less power-demanding, but higher-resolution and faster sensors. These allow for perceiving the finer details of objects, resulting in improved performance. In the following, we elaborate on high-resolution, multi-spectral, and 3D sensing devices used for horticultural robots.

\subsection{High-resolution sensing}

Stein \etal \cite{Stein2016-bg} exemplified mango detection and counting by exploiting 8.14M pixel images and a Light Detection And Ranging (LiDAR) system. High-resolution images enable detecting the details of plants, which leads to the extraction of useful and distinguishable features for fruit detection. A LiDAR generates an image mask of a tree by projecting 3D points of a segmented tree back to the camera plane for associating the fruit with the corresponding tree. This study reports impressive results in fruit counting (1.4\% over counting), and precision and accuracy ($R^2=0.90$).

\begin{figure}
\begin{center}
\includegraphics[height=0.38\columnwidth]{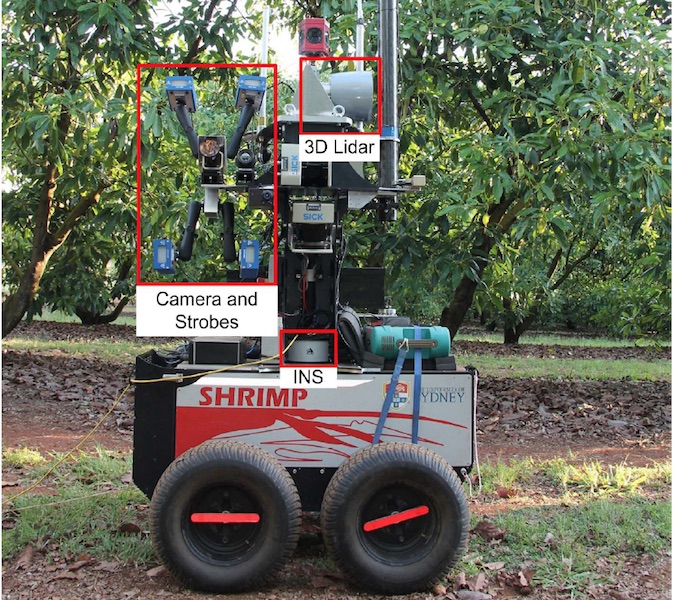}
\includegraphics[height=0.38\columnwidth]{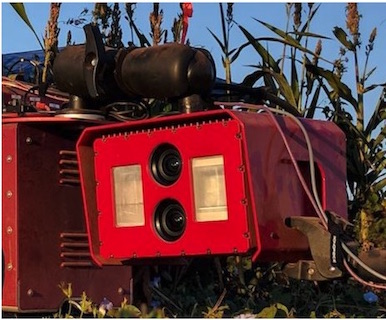}
\end{center}
    \vspace{-3mm}
    \caption{The robotics platform used in a mango orchard (left) \cite{Stein2016-bg} and a customized high-resolution stereo sensor used for imaging Sorghum stalks (right) \cite{baweja2018stalknet}.}
    \label{fig:high-res-cam}
    \vspace{-5mm}
\end{figure}

High-resolution cameras are not only useful for crop field scouting missions from the air, but also on the ground. Beweja \etal \cite{baweja2018stalknet} used a 9M pixel stereo-camera pair with a high-power flash triggered at 3\unit{Hz} (Fig.~\ref{fig:high-res-cam}). Stalks of Sorghum plants and their widths were detected and estimated for subsequent phenotyping. A disparity map calculated from the stereo images enables metric measurements with high precision (mean absolute error of 2.77\unit{mm}). They report their approach to be 30 and 270 times faster for counting and width measurements respectively compared to conventional manual methods. However, only a limited number of stalks were considered for counting (24) and width estimation (17).

\subsection{Multi-spectral sensing}
Detection performance can be improved by observing the infrared (IR) range in addition to the visible spectrum through multi-spectral (RGB+IR) images. In the past, this sensor was very costly due to the laborious manufacturing procedure behind multi-channel imagers, but they are now more affordable, with off-the-shelf commercial products readily available thanks to advances in sensing technologies. 

There are substantial studies on using multi-spectral cameras on harvesting and scouting robots in orchards \cite{hung2013orchard} and open fields (sweet peppers) \cite{mccool2016visual,bac2013robust}. The use of multi-spectral images improves classification performance by about 10\% with respect to RGB only models, with a global accuracy of 88\%. This improvement is analogous to that in \cite{mccool2016visual,bac2013robust} for detecting sweet peppers.

\subsection{From 2D to 3D: RGB-D sensing and LiDAR}
\begin{figure}
\begin{center}
\vspace{-4mm}
\subfloat{\includegraphics[height=0.36\columnwidth]{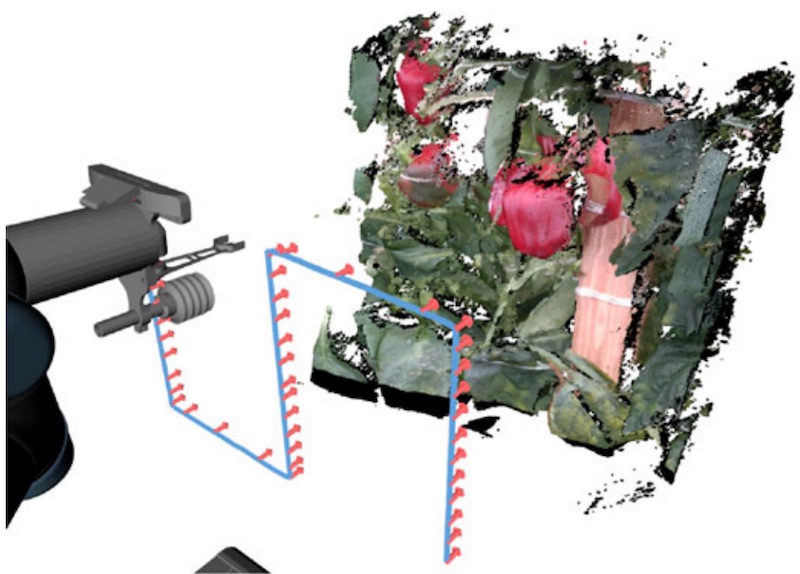}}
\subfloat{\includegraphics[height=0.36\columnwidth]{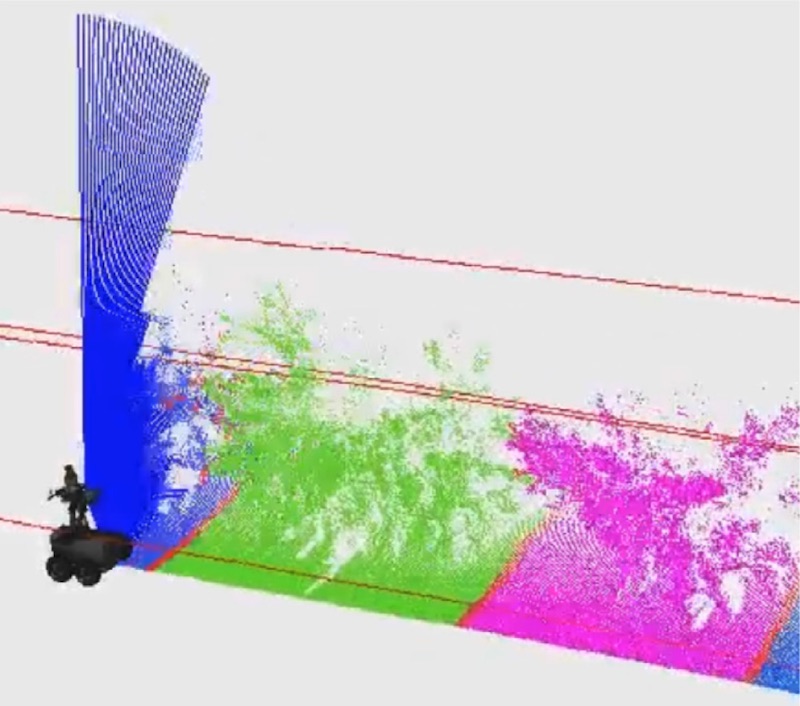}}

\end{center}
\vspace{-4mm}
	\caption{(Left) RGB-D sensing used for a protective sweet pepper farm reconstruction \cite{Lehnert:2016aa}. (Right) LiDAR mapping of almond trees in an orchard \cite{underwood2016mapping}.}
	\label{fig:3D-sensing}
	    \vspace{-3mm}
\end{figure}

Thus far, our work discussed the usage of 2D and passive sensing technologies for harvesting robots. However, advances in integrated circuit and microelectromechanical system (MEMs) technologies also unlock the potential of 3D sensing devices. For example, Red, Green, Blue, and Depth (RGB-D) sensing allows for constructing metric maps with high accuracy, as shown in Fig.~\ref{fig:3D-sensing}. This information is useful, not only for object detection, classification \cite{sa2017peduncle}, and fruit localization, but also for motion planning \cite{Lehnert:2016aa}, obstacle avoidance, and fine end-effector operation. A crucial step in the operation of RGB-D sensors is filtering out noise (de-noising) and outliers, which may be caused by poor sensor calibration, inaccurate disparity measurements due to ill-reflectance (e.g., under direct sunlight), etc., as highlighted in \cite{firman2016rgbd}. Using LiDAR is beneficial for longer-range scanning and mapping large fields \cite{Bargoti2017-kg,underwood2016mapping} (Fig.~\ref{fig:3D-sensing}). It is also necessary to design an RGB-D or LiDAR based harvesting system that can handle large amounts of incoming 3D data, which influences the cycle time of a harvesting robot, or the time for picking and placing a fruit. Table~\ref{tbl:sensor} summarizes our review of sensing technologies for harvesting robots.

\renewcommand{\tabcolsep}{1pt}
\begin{table}
\centering
\caption{Sensors for harvesting robots}
\vspace{-2mm}
\label{tbl:sensor}
\begin{threeparttable}
\begin{tabular}{cccccc}
\textbf{Sensors} & \textbf{\begin{tabular}[c]{@{}c@{}}Addressed \\ challenges\end{tabular}} & \textbf{Crops} & \textbf{\begin{tabular}[c]{@{}c@{}}Sensor \\ spec.\end{tabular}} & \textbf{Paper} & \textbf{Acc.}\tnote{a} \\  \midrule \midrule
\multirow{2}{*}{\begin{tabular}[c]{@{}c@{}}High-res \\ camera\end{tabular}} & \multirow{2}{*}{\begin{tabular}[c]{@{}c@{}}Deficiency \\ in detail\end{tabular}} & Mango & 8.14MP & \cite{Stein2016-bg} & 90\tnote{a} \\
 &  & Sorghum & 9MP & \cite{baweja2018stalknet} & 0.88\tnote{a}\\ \hline
\multirow{2}{*}{\begin{tabular}[c]{@{}c@{}}Multi-spectral \\ camera\end{tabular}} & \multirow{2}{*}{\begin{tabular}[c]{@{}c@{}}Improve \\ accuracy\end{tabular}} & Almond & RGB+IR & \cite{hung2013orchard} & 88\% \\
 &  & Sweet pepper & RGB+IR & \cite{mccool2016visual,bac2013robust} & 69.2\%, 58.9\%\\ \hline
\multirow{3}{*}{3D sensing} & \multirow{3}{*}{\begin{tabular}[c]{@{}c@{}}Metric,\\ motion \\ planning\end{tabular} } & Sweet pepper & RGB-D & \cite{Lehnert:2016aa,sa2017peduncle} & 80\%\tnote{b}, 0.71\tnote{c} \\
 &  & Almond trees & LiDAR & \cite{underwood2016mapping} & 0.77\tnote{a} \\
 &  & & & & \\ \hline
\end{tabular}

\begin{tablenotes}[para]
  \item[a] R-squared correlation.
  \item[b] Picking successful rate.
  \item[c] AUC of peduncle detection rate.
  \end{tablenotes}
  \end{threeparttable}
\vspace{-6mm}
\end{table}

%% file: contribs/Pollination2.tex
\section{Perception for pollination}\label{sec:Pollination2}


Early flower detection methods, e.g., in \cite{Das1999Color}, rely mainly on color values and do not perform well as many flowers have similar colors. To solve this issue, recent works consider additional information such as size, shape, and feature edges. Nilsback et al. \cite{Nilsback2006Color} used color and shape. Pornpanomchai et al. \cite{Pornpanomchai2010Herb} used RGB values with the flower size and the edge of petals feature to find herb flowers. Hong et al. \cite{Hong2012Color} found the contour of a flower image using both color- and edge-based contour detection. Tiay et al. \cite{Tiay2014Color} used a similar method, which uses edge and color characteristics of flower images. Yuan et al. \cite{Yuan2016TomatoPollination} used hue, saturation, and intensity (HSI) values to find flower, leaf and stem areas, before reducing noise with median and size filters. Bosch et al. \cite{Bosch2007Color} proposed image classification using random forests and compared multi-way Support Vector Machines (SVMs) with region of interest, visual word, and edge distributions. Kaur et al. \cite{Kaur2015Color} identified rose flowers using the Otsu algorithm and morphological operations. Bairwa et al. \cite{Bairwa2014Color} and Abinaya et al. \cite{Abinaya2016Color} proposed thresholding techniques to count gerbera and jasmine flowers, respectively. However, these color based approaches are not robust enough in variable lighting conditions.

To address this, recent research has considered deep learning techniques for flower detection. Yahata et al. \cite{Yahata2017CNN} proposed a hybrid image sensing method for flower phenotyping. Their pipeline is based on a coarse-to-fine approach, where candidate flower regions are first detected by Simple Linear Iterative Clustering (SLIC) and hue channel information, before the acceptance of flowers is decided by a convolutional neural network (CNN). Liu et al. \cite{Liu2016CNN} also developed a flower detection method based on CNNs. Srinivasan et al. \cite{Srinivasan2017CNN} developed a machine learning algorithm that receives a 2D RGB image and synthesizes an RGB-D light field (scene color and depth in each ray direction). It consists of a CNN that estimates scene geometry, a stage that renders a Lambertian light field using that geometry, and a second CNN that predicts occluded rays and non-Lambertian effects. While these approaches perform better compared to traditional methods, they demand high training times with large datasets on high-performance systems.

%% file: contribs/YieldEstimation2.tex
\section{Perception for Yield Estimation} \label{sec:Yield2}


Manual yield estimation is time-consuming, expensive, labor-intensive and inaccurate, with sometimes destructive outcomes. These aspects have motivated methods of process automation. However, fully automated estimation is a challenging task as: (a) the environment is unstructured and cluttered, (b) the fruit can have colors similar to the background, (c) they may lack distinguishable features and be occluded by other fruit, branches, or leaves, and above all, (d) there are uncontrolled illumination changes when yield estimation is done outdoors. In the following sub-sections, we elaborate on two different paradigms tackling these issues.

\subsection{Hand-crafted features}
Traditional yield estimation algorithms rely on visual detection methods using predefined hand-crafted features derived from image content. These can be based on various information, such as the shape, color, texture, or spatial orientation of the fruit using various feature representations such as the local binary pattern (LBP, texture features) \cite{Ojala2002-jt} and a histogram of gradients (HoG, geometry and structure features) \cite{Dalal2005-hd}, SIFT \cite{Lowe2004-wv} or SURF\cite{Bay2006-hd} (key-points features). For example, Nuske et al.~\cite{Nuske2011YieldEI} and Li et al.~\cite{Li2017GreenAR} utilized shape and texture features to detect grapes and green apples, respectively. Wang et al.~\cite{Wang2012AutomatedCY} and Linker~\cite{Linker2016APF} exploited color and specular reflection to detect apples. Verma et al.~\cite{Verma2014SegmentationOT} and Dorj et al.~\cite{Dorj2013ACV} used color-space features to detect tomatoes and tangerines, respectively.

The recent survey by Gongal et al.~\cite{Gongal2015SensorsAS} reviews computer vision for fruit detection and localization. This paper draws the following conclusions: 
\begin{itemize}
    \item learning-based methods are superior to simple threshold-based image segmentation methods for fruit detection in realistic environments,
    \item combining multiple types of hand-crafted features is better than using only one type of feature,
    \item detection methods based on hand-crafted features perform poorly when faced with occlusions, overlapping fruit and variable lighting conditions.
\end{itemize}

After the above survey was published, there was a major breakthrough in object detection and localization using Deep Neural Networks (DNNs) for learned feature extraction. In the next section, we examine the new shift towards deep learning for yield estimation applications.

\subsection{Learning-based features}
\begin{figure}
\begin{center}
\includegraphics[width=\columnwidth]{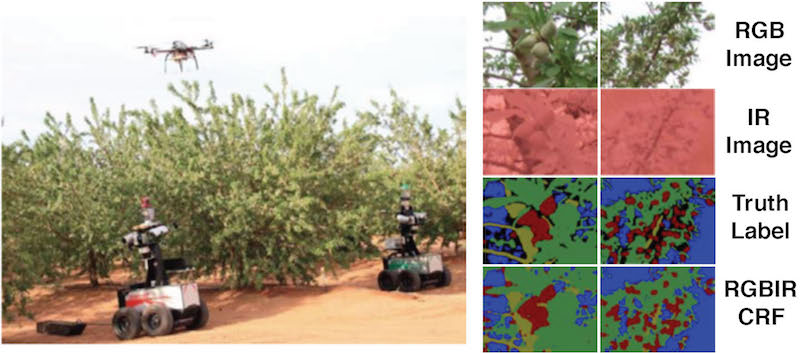}
\end{center}
    \vspace{-3mm}
    \caption{Orchard almond yield estimation using multi-spectral images \cite{hung2013orchard}.}
    \label{fig:almondUAV}
    \vspace{-5mm}
\end{figure}

One of the earliest works using learned features was applied to segment almond fruit~\cite{hung2013orchard} (Fig.~\ref{fig:almondUAV}). Visual features from multi-spectral images were learned using a sparse autoencoder at different image scales, followed by a logistic regression classifier to learn pixel label associations. 

Results show that leveraging a learning approach for feature extraction renders the system more robust to illumination changes. Stein et al.~\cite{Stein2016ImageBM} proposed a mango fruit detection and tracking system based on a faster region-based Convolutional Neural Network (Faster R-CNN)~\cite{Ren2015FasterRT} using detection and camera trajectory information to establish pair-wise correspondences between consecutive images. 

Rahnemoonfar and Sheppard~\cite{Rahnemoonfar2017DeepCF} proposed a fruit counting system that employs a modified version of the Inception-ResNet architecture~\cite{Szegedy2017Inceptionv4IA} trained on synthetic data. The network predicts the number of fruit from the input image directly, without segmentation or object detection as intermediate steps. Recently, Halstead et al.~\cite{Halstead2018FruitQA} proposed a sweet pepper detection and tracking system inspired by the DeepFruits detector~\cite{sa2016deepfruits}. 
The system is trained to perform efficient in-field assessment of both fruit quantity and quality.

Despite the high accuracy measures reported in the works above, using deep learning for fruit detection still faces the following challenges:
(a) It requires vast amounts of labeled data, which is time-consuming and can be expensive to obtain. Using synthetic data from generative models, e.g., in \cite{Rahnemoonfar2017DeepCF} and \cite{Barth2018}, has recently emerged as a method of addressing this issue.
(b) Tracking and data association is crucial to prevent over-counting produce. Not all works reviewed address this aspect, as they estimate yield from only a single image of a tree such that only fruit in the current view are counted. Manual calibration is usually carried out to infer the total amount of fruit based on the visible proportion. However, this process is required for every tree species, and also varies between years.
(c) There is a lack of independent third-party benchmark tests for estimating yields for various fruit. Currently, results are reported based on in-house datasets, which can be very small, making reported results hard to compare fairly.

%% file: contribs/Harvesting2.tex
\section{Perception for Harvesting}\label{sec:harvesting2}
Traditionally, harvesting robots exploit traditional hand-crafted approaches (e.g., extracting useful visual or geometric features) \cite{sa2017peduncle,bac2013robust} and plantation geometries (e.g., tree rows in orchards) \cite{underwood2016mapping} for crop perception. While these methods show promising results, our survey concentrates on the new paradigm of using data-driven DNNs. The variety of off-the-shelf DNNs today with human-level performance \cite{geirhos2017comparing,eckstein2017humans} indicates they are transitioning from dataset benchmarking into in-situ production environments. In this section, we investigate strategies for tackling the major perceptual challenges in automated harvesting: occlusions, perception aliasing, and environmental variability.

\subsection{Occlusions}
\begin{figure}
\begin{center}
\vspace{-2mm}
\subfloat{\includegraphics[height=0.3\columnwidth]{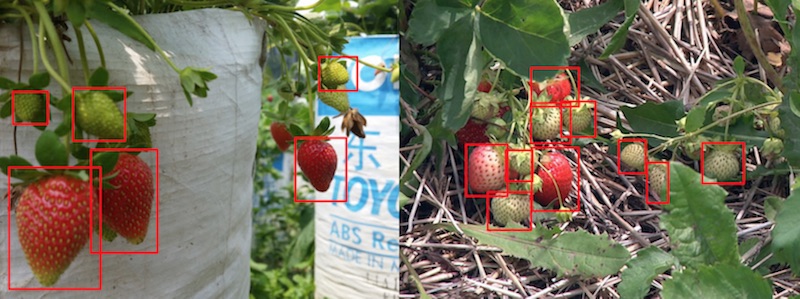}}\\ \vspace{-3mm}
\subfloat{\includegraphics[height=0.3\columnwidth]{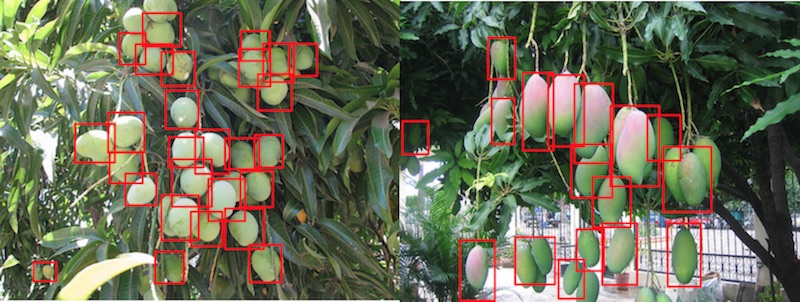}}

\end{center}
\vspace{-4mm}
	\caption{Examples of detecting (top) strawberry and (bottom) mango fruits using the DeepFruits network\cite{sa2016deepfruits}.}
	\label{fig:deepFruits}
	    \vspace{-3mm}
\end{figure}

Harvesting scenes commonly exhibit occlusions of crops caused by themselves or other plant parts (leafs, stems or peduncles). As a result, it may be difficult to detect crops using only low-level features, such as colour, texture, and shape, etc., and instead necessary to employ DNNs for higher-level contextual scene understanding through convolutional multi layers. 
To this end, typically a large number of internal parameters (e.g., weights and biases) must be properly tuned, which is a nontrivial task. This training process requires a vast number of samples to avoid over-fitting to small datasets. It is thus common practice to use pre-trained parameters fitted on millions of samples (e.g., images) for variable initialization; to fine-tune. After fine-tuning, the trained DNN can be refined with relatively small datasets.
Sa \etal \cite{sa2016deepfruits} exemplified this procedure, as shown in Fig.~\ref{fig:deepFruits}. 602 images of seven fruit were considered for model training and testing with $\sim$0.9 F1-score for most fruit. Although the trained model handles occlusions reasonably, it struggles with large variations in training and testing images as it expects visually similar environments. 

\subsection{Homogeneous farm fields: perception aliasing}
It is common practice to seed crops and trees in linear row arrangements separated by vacant space. This geometric formation can be useful for perception by providing an informed prior for plant detection \cite{lottes2017uav}. However, such a structure also creates the issue of perception aliasing, which particularly impedes accurate mapping and localization in farm fields. Many crops resembling each other in appearance produce high false positive rates, which degrades harvesting performance. Recently, Kraemer \etal \cite{kraemer2017plants} proposed a method of creating landmarks from plants using an FCNN. These landmarks can be used for robot pose localization and mapping farm environments. It is also possible to address this issue by fusing different sensing modalities, e.g., from an Inertial Measurement Unit (IMU), wheel odometry, LiDAR, or Global Positioning System (GPS), to bound a search space for visual matching.
    
\subsection{Environmental variability}
Harvesting robots usually operate in a wide range of perceptual conditions, which may feature various lightning conditions, dynamic objects, and unbounded crop scales. Researchers have attempted to restrict operating environments using light-controllable greenhouses or by operating at night time with high-power artificial flashes rather than harvesting in open fields. While these constraints increase production costs and reduce operation time, they do improve perceptual performance. Chen \etal \cite{chen2017counting} demonstrated an approach to detect and count oranges and apples with data-driven FCNN. A mean intersection of union of 0.813 and 0.838 were achieved for oranges and apples, respectively. Table~\ref{tbl:perception-harvesting} summarizes our study of perception for harvesting robots.

\renewcommand{\tabcolsep}{6pt}
\begin{table}[]
\centering
\caption{Perception trend of harvesting robots}
\vspace{-2mm}
\label{tbl:perception-harvesting}
\begin{threeparttable}
\begin{tabular}{cccccc}
\textbf{Paper}                      & \begin{tabular}[c]{@{}c@{}}\textbf{Network}\\ \textbf{arch.}\end{tabular} & \textbf{Crops}                                                   & \begin{tabular}[c]{@{}c@{}}\textbf{Classification}\\ \textbf{type}\end{tabular}    & \begin{tabular}[c]{@{}c@{}}\ \textbf{\# images}\\ \textbf{(tr/test)}\end{tabular} & \textbf{Accu.} \\ \hline \hline
\cite{sa2016deepfruits}  & \begin{tabular}[c]{@{}c@{}}Faster \\ R-CNN\end{tabular} & 7 fruits                                                & Obj. based                                                     & \begin{tabular}[c]{@{}c@{}}484/118\\ RGB\end{tabular}         & 0.9\tnote{a}    \\ \hline
\cite{kraemer2017plants} & FCNN                                                    & \begin{tabular}[c]{@{}c@{}}Root\\ crops\end{tabular}    & Pixel based                                                    & \begin{tabular}[c]{@{}c@{}}1398/300\\ RGB+IR\end{tabular}     & 0.9\tnote{b}    \\ \hline
\cite{chen2017counting}  & FCNN                                                    & \begin{tabular}[c]{@{}c@{}}Orchard\\ fruits\end{tabular} & \begin{tabular}[c]{@{}c@{}}Pixel and\\ obj. based\end{tabular} & \begin{tabular}[c]{@{}c@{}}47/45\\ RGB\end{tabular}           & 0.838\tnote{c}  \\ \hline
\cite{sa2018weednet}     & Segnet                                                  & \begin{tabular}[c]{@{}c@{}}Root\\ crops\end{tabular}    & Pixel based                                                    & \begin{tabular}[c]{@{}c@{}}375/90\\ RGB+IR\end{tabular}       & 0.8\tnote{a}    \\ \hline
\end{tabular}
\begin{tablenotes}[para]
  \item[a] F1-score 
  \item[b] precision/recall rate
  \item[c] mean intersection of union
  \end{tablenotes}
  \end{threeparttable}
\vspace{-3mm}
\end{table}

%% file: contribs/Conclusions2.tex
\section{Conclusions}\label{sec:conclusion2}


To the best of the authors' knowledge, this survey is the first of its kind to cope with the three major inter-connected horticultural procedures of pollination, yield estimation, and harvesting in the context of autonomous robotics. 
Our discussions in this work target practical challenges in perception that inhibit robot deployment in real production scenarios. We tried to supplement, rather than replicate, current survey papers by focusing on most contemporary works not covered by these reviews.

Our survey exposed that the main challenges facing the development of automated pollination robots involve selecting hardware and sensing equipment, robust flower detection, and row-following on uneven and bumpy surfaces. Whereas traditional hand-crafted features with classifiers have been widely exploited for flower detection, there is a rapid paradigm shift towards DNNs. Complementary multi-modal sensing is an essential element for robust vehicle navigation to compensate for uneven outdoor environments.

In automated yield estimation, our survey revealed challenges in crop detection in unstructured and cluttered environments, uncontrolled lighting conditions, and occlusions caused by leaves, branches, and other crops. Here, machine learning also plays a pivotal role in crop detection and localization, and DNNs pave the way for overcoming occlusions and illumination changes.

Issues in automated harvesting closely resemble those in yield estimation, with additional difficulties arising in designing manipulators and end-effectors. Unless there are special requirements, it is desirable to use commercial manipulators to minimize development time and effort, with only a custom and application-specific end-effector design. Exploiting geometrical prior knowledge about fields, such as crop rows, to improve performance is viable for common horticultural practices.

While our review of recent developments in DNNs and GPU-driven computing uncovers their potential in horticulture, several open challenges remain. Namely, the state-of-art requires larger, more accessible datasets to prevent cases of model over-fitting, as well as faster processing devices to enable real field deployments.

%% file: contribs/Acknowledgement.tex
\section*{Acknowledgement}\label{sec:ack}
\vspace{-1mm}
This study has received support from the European Union's Horizon 2020 research and innovation programme under grant agreement No 644227, (Flourish) from the Swiss State Secretariat for Education, Research and Innovation (SERI) under contract number 15.0029. It was also supported by the New Zealand Ministry for Business, Innovation and Employment (MBIE) on contract UOAX1414. We would like to thank to New Zealand MBIE Orchard project team (The University of Auckland, Plant and Food Research, The University of Waikato, Robotics Plus).